# When Gender is Hard to See: Multi-Attribute Support for Long-Range Recognition

Nzakiese Mbongo[a,c], Kailash A. Hambarde[a,b], Hugo Proença[a,b]

[a] *University of Beira Interior, Portugal*
[b] *IT: Instituto de Telecomunicações, Portugal*
[c] *Institute of Information and Communication Technologies, University of Luanda, Angola*

**Abstract**

Accurate gender recognition from extreme long-range imagery remains a challenging problem due to limited spatial resolution, viewpoint variability, and loss of facial cues. For such purpose, we present a dual-path transformer framework that leverages CLIP to jointly model visual and attribute-driven cues for gender recognition at a distance. The framework integrates two complementary streams: (1) a direct visual path that refines a pre-trained CLIP image encoder through selective fine-tuning of its upper layers, and (2) an attribute-mediated path that infers gender from a set of soft-biometric prompts (e.g., hairstyle, clothing, accessories) aligned in the CLIP text–image space. Spatial–channel attention modules further enhance discriminative localization under occlusion and low resolution. To support large-scale evaluation, we construct U-DetAGReID, a unified long-range gender dataset derived from DetReIDx and AG-ReID.v2, harmonized under a consistent ternary labeling scheme (Male, Female, Unknown). Extensive experiments suggest that the proposed solution surpasses state-of-the-art person-attribute and re-identification baselines across multiple metrics (macro-F1, accuracy, AUC), with consistent robustness to distance, angle, and height variations. Qualitative attention visualizations confirm interpretable attribute localization and responsible abstention behavior. Our results show that language-guided dual-path learning offers a principled, extensible foundation for responsible gender recognition in unconstrained long-range scenarios.https://github.com/Villjoie/LMASD-G.

## 1. Introduction

Recognizing pedestrians gender from extreme long-range imagery (up to 120 m) captured by drones or surveillance cameras is highly challenging. At such distances, faces are blurred, occluded, or entirely missing, and fine-grained body cues become unreliable due to low resolution, motion blur, or oblique viewpoints [11, 25, 1]. Yet, gender perception at scale remains valuable for demographic analytics and crowd understanding.

Traditional approaches rely on facial or close-up body features, which collapse under these constraints. We instead explore soft biometrics [34, 4]—contextual attributes such as hairstyle, clothing, footwear, and accessories—that, while individually ambiguous, collectively provide meaningful evidence. Importantly, Gender perception often emerges from the combination of multiple weak signals rather than any single feature. Existing binary classifiers fail in such scenarios, both ethically and statistically, by forcing uncertain predictions. To counter this, we explicitly introduce an "*Unknown*" class to allow responsible abstention when evidence is insufficient.

We propose a dual-path multimodal architecture that fuses direct visual evidence and attribute-mediated reason-ing via CLIP-based vision–language alignment. The model aggregates per-attribute text→vision attentions into a unified semantic representation (Fig. 1). A spatial–channel attention (SCA) module enhances discriminative localization, while selective fine-tuning of the last CLIP blocks preserves pre-trained priors. Text prompts (e.g., emph"a person wearing high heels") act as semantic queries that steer attention toward relevant regions, enabling interpretable and robust inference across viewpoints and distances.

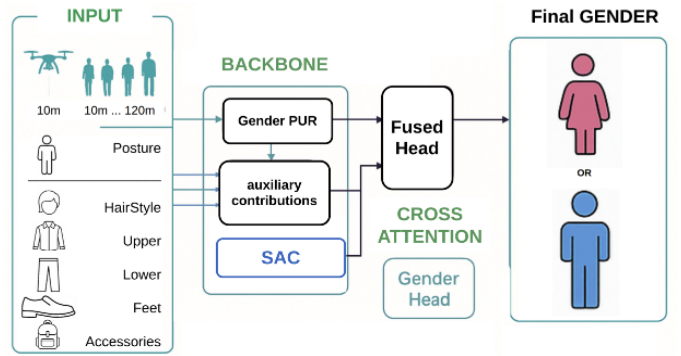

Figure 1: Cohesive view: attribute contribution maps derived from text-conditioned *Transformer attention rollout* [31]. Each map highlights the region most correlated with a given attribute prompt (*Hairstyle*, *Upper*, *Lower*, *Feet*, *Accessories*), and the fused *Gender* output.

This design yields three concrete benefits: (1) Robustness under missing or degraded cues; (2) Explainability via interpretable attention at both attribute and gender levels; (3) Stronger generalization across sensors, datasets, and environments.

To evaluate long-range robustness, we introduce U-DetAGReID, a unified benchmark combining DetReIDx and AG-ReID.v2 under a consistent ternary labeling scheme (*Male, Female, Unknown*). According to our experiments, our solution achieves consistent gains in macro-F1, accuracy, and AUC across distance, angle, and height strata, surpassing strong baselines. Qualitative analysis confirms attribute-aware attention and reliable abstention behavior under uncertainty.

**Contributions.** Our key contributions are: (i) a dual-path CLIP-based framework integrating direct and attribute-mediated gender reasoning; (ii) semantic composition of soft-biometric evidence through text→vision attention; (iii) selective fine-tuning with spatial–channel attention for efficient long-range adaptation; (iv) a unified dataset protocol (U-DetAGReID) with ternary labels for responsible evaluation; and (v) comprehensive ablations and explainability analyses validating each design choice.

## 2. Related Work

*Classical PAR and imbalance-aware losses..* Early PAR evolved from handcrafted descriptors to deep CNNs. *DeepMAR* [26] introduced joint multi-label learning with *weighted sigmoid cross-entropy* to tackle severe imbalance, improving PETA by exploiting label correlations (e.g., *skirt–female*). Yet, reliance on global representations and limited spatial context harms robustness under occlusion and long-range. *Label-balanced Multi-label Learning (LML)* [18] decouples attribute-wise re-sampling from co-occurrence and adds Bayesian feature augmentation for rare attributes, but still assumes balanced, close-range imagery—less suited to aerial/drone PAR.

*Weak localization and attention mechanisms..* Weakly supervised localization aligns attributes with regions via attention. *ALM* [20] learns multi-scale attribute–region correspondences; *VAC* [21] enforces attention consistency under geometric transforms, stabilizing multi-label predictions. Despite gains, most works presume visible, near-frontal pedestrians; when parts occupy only a few pixels (hair, feet, bag), attention degrades. We extend this line with *text→vision* attention in CLIP's space, guiding semantic region selection even in distant aerial views.

*Long-range dependencies and efficiency..* To capture broader context, recent models introduce long-range mechanisms. *MambaPAR* [17] employs Mamba blocks for spatial–semantic interaction; *SNN-PAR* [22] leverages spiking networks with dual-level distillation for energy efficiency. These advances, however, largely assume ground-level detail. We instead target extreme long range (~120 m) with dual CLIP encoders and selective fine-tuning of late visual layers to retain pretrained knowledge while adapting to aerial/oblique viewpoints.

*Reliable evaluation and unified protocols..* Rethinking PAR [19] exposed large performance variance across datasets and losses, advocating unified training and identity-independent splits. Following this, we use a single protocol (input size, optimizer, schedule) across baselines and ablations for fair comparison.

*Attributes + Re-ID and correlation modeling..* Multi-task designs such as *APR Network* [23] jointly learn identity and attributes via an *Attribute Re-weighting Module*, leveraging inter-attribute dependencies to improve both tasks. Rather than explicit label graphs, we compose evidence through prompts and cross-attention between a direct visual path and an attribute-mediated hypothesis.

*CLIP and multimodal generalization..* Vision–language formulations (e.g., *AttriVision*) show CLIP embeddings paired with focal losses can improve generalization on unified PAR benchmarks. Distinct from single-branch CLIP baselines, we employ two complementary CLIP paths—a purely visual stream and a text-conditioned attribute stream—fused via cross-attention into a ternary decision (Male/Female/Unknown), addressing ambiguity and missing facial cues typical of aerial footage.

## 3. Proposed Methodology

Our method leverages CLIP's vision–language space to integrate *soft biometrics* via textual *prompts*. The architecture has two parallel paths: a direct visual path for global gender cues and an attribute-mediated path for semantic composition; their hypotheses are fused via cross-attention for the final prediction. We assume a batch $\mathbf{X} \in \mathbb{R}^{B \times 3 \times H \times W}$ ($H = W = 224$) and prompts derived from attribute descriptions.

### 3.1. Overview

We adopt a multimodal two-path design with attribute composition. Path 1 yields a *direct visual* gender hypothesis; Path 2 builds text-guided per-attribute representations and an *attribute-mediated* hypothesis. The two are merged by *cross-attention* (§3.2.5) to produce the final decision. Both paths can include spatial–channel attention (SCA) to enhance salient visual tokens. Visual features come from CLIP ViT-B/16 at 224×224 (positional interpolation) [7], preserving the joint 512-D alignment.

We use two CLIP ViT-B/16 encoders [7, 8], one per path. We select five long-range "champion" soft biometrics—*hairstyle, lower, feet, accessories, upper*—[4] instantiated from a curated vocabulary (see §4), with class counts $K_a$ for each $a \in \mathcal{A}$.

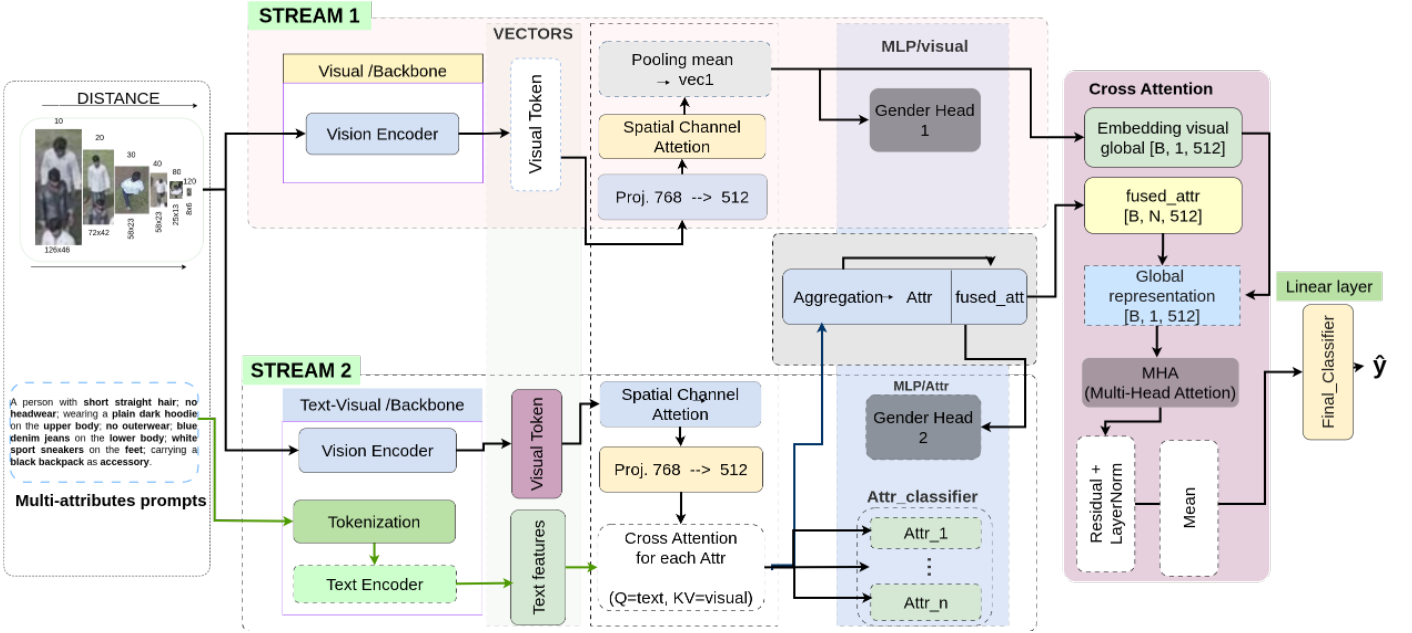

Figure 2: Overview of the proposed solution. Stream 1 uses CLIP ViT-B/16 visual tokens, projected (768→512), optionally refined by Spatial Channel Attention (SCA), mean-pooled into **vec1** and passed to an auxiliary gender MLP (Gender Head 1). Stream 2 encodes multi-attribute prompts with the CLIP text encoder; visual tokens are projected and refined, then a text-to-vision multi-head attention (Q=text, K/V=visual) yields per-attribute vectors $r^a$ for attribute classification. These vectors are aggregated into **fused_attr** and fed to a second auxiliary gender MLP (Gender Head 2). For fusion, **vec1** and **fused_attr** form a two-token sequence processed by a multi-head attention block with residual connection and LayerNorm; the outputs are averaged and passed through linear layers to obtain the final three-class gender prediction.

In *Path 1*, we aggregate tokens into a global embedding $\mathbf{v}_1 \in \mathbb{R}^{B \times 512}$ to produce the direct gender logits $\hat{\mathbf{g}}_1 \in \mathbb{R}^{B \times 3}$ (Male= 0, Female= 1, Unknown= 2). In *Path 2*, text→vision attention localizes per-attribute regions $\{\mathbf{r}^a\}_{a \in \mathcal{A}} \in \mathbb{R}^{B \times 512}$, yielding attribute logits and an aggregated embedding $\mathbf{v}_2 \in \mathbb{R}^{B \times 512}$ for the auxiliary hypothesis $\hat{\mathbf{g}}_2$. Fusion combines $\mathbf{v}_1$ and $\mathbf{v}_2$ into $\mathbf{v}_f \in \mathbb{R}^{B \times 512}$, producing the final output $\hat{\mathbf{g}}_f \in \mathbb{R}^{B \times 3}$.

## 3.2. Architecture

### 3.2.1. Token Extraction and Projection to 512-D

Images are resized to $224 \times 224$ and normalized with CLIP statistics (mean $\mu = [0.4815, 0.4578, 0.4082]$, std $\sigma = [0.2686, 0.2613, 0.2758]$). The CLIP visual encoder outputs a grid of tokens (without [CLS]) with $d_v = 768$ and $N = 21^2 = 441$ patches. Each token is projected into CLIP's multimodal space $d = 512$:

$$\mathbf{Z} = \mathbf{T}\,\mathbf{W}_p + \mathbf{b}_p \ \in \ \mathbb{R}^{B \times N \times 512}. \quad (1)$$

### 3.2.2. Spatial–Channel Attention (SCA)

We reshape $\mathbf{Z}$ into a feature map $\mathbf{F} \in \mathbb{R}^{B \times 512 \times 21 \times 21}$ and apply SE (channel) followed by CBAM (spatial) [32], with residual skip:

$$\mathbf{F}' = \underbrace{\mathbf{F} \odot \mathbf{A}_c(\mathbf{F})}_{\text{channel}} \odot \underbrace{\sigma\bigl(\mathrm{Conv}_{7\times 7}([\mathrm{Max}_c(\cdot) \,\|\, \mathrm{Avg}_c(\cdot)])\bigr)}_{\text{spatial}} + \mathbf{F}, \quad (2)$$

then reshape back to tokens $\mathbf{Z}' = \mathrm{Reshape}^{-1}(\mathbf{F}')$. SCA can be enabled independently in each path (flags `use_sca_path1/2`).

### 3.2.3. Path 1 — Direct (Visual) Gender

We globally average the tokens (after SCA, if enabled):

$$\mathbf{v}_1 = \tfrac{1}{N}\sum_{i=1}^{N} \mathbf{Z}'_{[:,i,:]} \ \in \ \mathbb{R}^{B \times 512}, \quad (3)$$

and feed a shallow MLP head to obtain $\hat{\mathbf{g}}_1 \in \mathbb{R}^{B \times 3}$.

### 3.2.4. Path 2 — Attributes (Text→Vision) and Mediated Gender

For each $a \in \mathcal{A}$, a short prompt [32] is encoded by CLIP's text encoder, producing $\mathbf{q}^a \in \mathbb{R}^{B \times 512}$. We apply pure MultiheadAttention (no LN/residual at this point) with $\mathbf{Q} = \mathbf{q}^a$, $\mathbf{K} = \mathbf{Z}'$, $\mathbf{V} = \mathbf{Z}'$ [33]:

$$\mathbf{R}^a = \mathrm{MHA}(\mathbf{q}^a,\ \mathbf{Z}',\ \mathbf{Z}') \in \mathbb{R}^{B \times 1 \times 512}, \quad \mathbf{r}^a = \mathbf{R}^a_{[:,1,:]}. \quad (4)$$

Each $\mathbf{r}^a$ feeds a linear multi-class head $\hat{\mathbf{y}}^a \in \mathbb{R}^{B \times K_a}$. We aggregate regions as

$$\mathbf{v}_2 = \tfrac{1}{|\mathcal{A}|}\sum_{a \in \mathcal{A}} \mathbf{r}^a \ \in \ \mathbb{R}^{B \times 512}, \quad (5)$$

and produce the mediated hypothesis $\hat{\mathbf{g}}_2 \in \mathbb{R}^{B \times 3}$. In the dataset, we concatenate short, focused prompts per sample (sourced from the CSV) prior to CLIP tokenization, keeping phrases concise.

### 3.2.5. Cross-Attention Fusion

We stack the two hypotheses as tokens $\mathbf{U} = [\mathbf{v}_1; \mathbf{v}_2] \in \mathbb{R}^{B \times 2 \times 512}$ and apply MHA + residual + LayerNorm:

$$\mathbf{H} = \text{MHA}(\mathbf{U}, \mathbf{U}, \mathbf{U}), \quad \tilde{\mathbf{H}} = \text{LN}(\mathbf{H} + \mathbf{U}), \quad (6)$$

followed by mean pooling over tokens and a linear projection to obtain the final logits $\hat{\mathbf{g}}_f \in \mathbb{R}^{B \times 3}$.

### 3.3. Semantic Composition and Explainability

The final decision is informed by semantic composition of attributes: prompt-guided regions make explicit the cues supporting the prediction, approximating human reasoning when faces are not visible. The model affords explainability via per-attribute scores and text→vision attention maps.

### 3.4. Loss Function

*Three-class gender with auxiliary supervision..* Let $\mathbf{y}_g \in \{0, 1, 2\}^B$. We use categorical cross-entropy on the final output and, with auxiliary weight $\alpha = 0.25$, on both paths:

$$\mathcal{L}_g = \text{CE}(\hat{\mathbf{g}}_f, \mathbf{y}_g) + \alpha \Big[ \text{CE}(\hat{\mathbf{g}}_1, \mathbf{y}_g) + \text{CE}(\hat{\mathbf{g}}_2, \mathbf{y}_g) \Big]. \quad (7)$$

*Attributes (per-attribute multiclass)..* For each $a \in \mathcal{A}$ with $K_a$ classes and an ignore index for missing labels:

$$\mathcal{L}_a = \frac{1}{|\mathcal{A}|} \sum_{a \in \mathcal{A}} \text{CE}(\hat{\mathbf{y}}^a, \mathbf{y}^a), \quad (8)$$

and the total loss is

$$\mathcal{L}_{\text{total}} = \lambda_g \mathcal{L}_g + \lambda_a \mathcal{L}_a, \quad \lambda_g = \lambda_a = 0.5. \quad (9)$$

(We contrast this with FCE losses used in recent PAR works; here we favor supervised discriminative heads in CLIP's 512-D space.)

### 3.5. Optimization and Selective Fine-Tuning

We train all new modules (projections, SCA, heads) end-to-end and apply selective fine-tuning to CLIP, unfreezing only the last $L_v = 4$ visual blocks by default (textual layers optional). We use differential learning rates—lower for CLIP, higher for newly added modules—together with standard scheduling and gradient clipping. Concrete hyperparameters and the unfreezing policy are detailed in §4. The model employs two independent CLIP ViT-B/16 encoders (12 visual blocks each).

### 3.6. Data Pipeline, Inference, and Metrics

Pipeline. Images are resized to 224×224 and normalized with CLIP statistics. Training uses random horizontal flip; validation uses resize+normalization only. The loader returns the image, three-class gender (Male/Female/Unknown), per-attribute labels (ignore index −100), and the aggregated per-sample prompt.

Inference. The model consumes the image and attribute text (from the vocabulary or standardized prompts) and outputs $\hat{\mathbf{g}}_1, \hat{\mathbf{g}}_2, \hat{\mathbf{g}}_f$ plus per-attribute logits.

Metrics & Visualizations. We report *accuracy*, *balanced accuracy*, macro *F1/Precision*, weighted *Recall*, AUC (Female vs Rest), and confusion matrices, alongside text→vision attention visualizations. Section 4 includes analyses by capture distance and camera elevation.

## 4. Experiments and Results

### 4.1. Datasets and Label Harmonization

We use two complementary, attribute-annotated datasets: **DetReIDx** and **AG-ReID.v2**. Both cover real-world scenes (urban/campus) with multi-camera captures, including aerial (drone) views; our unified corpus **U-DetAGReID** focuses on aerial imagery. Annotations span 15 soft-biometrics: gender (three-class: male/female/unknown), age, height, weight, ethnicity, hair color, hairstyle, beard, mustache, glasses, head (coverage/hat), upper (torso), lower (legs), feet (footwear), and bag/accessories.

To train consistently across sources, we build **U-DetAGReID** by harmonizing both ontologies and standardizing class indices per attribute, *explicitly preserving the Unknown class* where evidence is insufficient. This unified label space enables consistent training and evaluation on heterogeneous data.

*Target attributes and gender label..* From the 15 attributes, we select the most discriminative and long-range visible cues—*hairstyle*, *upper*, *lower*, *feet*, *accessories/bag*—and train jointly with the three-class gender label {Male, Female, Unknown}. Retaining *Unknown* prevents forced decisions under low resolution, occlusion, or unfavorable viewpoints.

*Ontology unification..* We map native categories from each dataset to a common per-attribute ontology, avoiding index collisions. Classes absent in a source remain unassigned; uncertain annotations are kept as *Unknown*. For brevity we omit per-attribute tables here. All mapping specs, CSVs, and build scripts for **U-DetAGReID** are available at: https://github.com/Villjoie/LMASD-G.

*Prompt vocabulary..* Text→vision prompts come from a CSV with expanded descriptions per attribute/class (e.g., *feet*: "a person wearing high heels"). For each sample, active descriptions are concatenated into a short sentence and tokenized by CLIP's text encoder; this acts as the textual query driving text→vision attention and aligning regions with the curated vocabulary. For missing/undefined labels we use neutral prompts (e.g., "the attribute is unclear") or omit terms to avoid bias.

Table 1: **Dataset partitioning.** Counts by gender and totals in **U-DetAGReID**.

| Dataset | Train | | | Val | | | Test | | | Total Samples |
|---|---|---|---|---|---|---|---|---|---|---|
| | M | F | Total | M | F | Total | M | F | Total | |
| U-DetAGReID | 177 173 | 131 419 | 308 592 | 110 555 | 84 128 | 194 683 | 67 591 | 67 361 | 134 952 | 638 372 |

Split totals computed as M+F. Global total = 308,592 + 194,683 + 134,952 = 638,227.

*Splits and preprocessing..* We keep each dataset's official splits and ensure no identity leakage across train/val/test in the unified corpus. Images are resized to 224×224 and normalized with CLIP statistics; training uses random horizontal flip, validation uses resize+normalization only. When applicable, the camera/scene distribution is preserved. The final **U-DetAGReID** partition is shown in Table 1.

*Note on distance/height..* To analyze performance versus capture distance and camera elevation, we leverage dataset metadata and distributions from **AG-ReID.v2** [14, 25] and **DetReIDx** [11, 15, 16]. When direct metadata is unavailable, we use measurable proxies—e.g., pedestrian apparent resolution (bounding-box height, in pixels) and relative scale—and stratify samples into predefined bins. In the unified **U-DetAGReID**, capture distances span 10–120 m, elevations 5.8–90 m, and Euclidean camera–target distances up to 150 m. Results are reported per bin of these quantities to expose the gradual degradation as distance/elevation increases.

*4.2. Implementation Details*

We summarize the settings used to reproduce our solution on **U-DetAGReID**. All code follows standard PyTorch practice with selective CLIP fine-tuning to balance pretraining preservation and domain adaptation.

*Training protocol..* 60 epochs, mini-batch 32. Optimizer: Adam with ($\beta_1 = 0.9$, $\beta_2 = 0.999$) and weight decay $1.5\times 10^{-4}$. Two LR groups: $\eta_{\text{CLIP}} = 10^{-6}$ for unfrozen CLIP layers and $\eta_{\text{HEADS}} = 10^{-4}$ for projections, SCA modules, and classifier heads. Stepwise decay (×0.1) at epochs {20, 40}. Gradient clipping: global norm 5.0. Mixed precision (AMP) enabled.

*Selective fine-tuning..* Two independent CLIP ViT-B/16 encoders (12 visual blocks each). We unfreeze only the last $L_v = 4$ visual blocks per encoder; textual layers remain frozen ($L_t = 0$). All newly introduced modules (projections, SCA, heads) are trained end-to-end.

*Preprocessing and augmentation..* Inputs resized to 224×224 and normalized with CLIP mean/std. Training uses random horizontal flip; validation uses resize+normalization only.

Table 2: **Key implementation settings.**

| Category | Setting |
|---|---|
| Framework | PyTorch; HuggingFace `CLIPModel`; codebase `MultiModalGenderClassifier`. |
| Backbone | Two CLIP ViT-B/16 encoders (`clip1`, `clip2`). |
| Inputs | 224×224; CLIP stats $\mu = [0.4815, 0.4578, 0.4082]$, $\sigma = [0.2686, 0.2613, 0.2758]$. |
| Attrs | `{hairstyle, lower, feet, accessories, upper}` (CSV prompts). |
| Heads | Gender: *direct*, *attribute-mediated*, *fused* (3-class). |
| Fusion | Cross-attention over `vec1` and `fused_attr`; SCA optional per path ($r$=16). |
| Data | Train: random horizontal flip; Val: resize+normalization; fixed attribute order when 7 attrs are used. |
| Opt. | AdamW ($\beta_1$=0.9, $\beta_2$=0.999), weight decay $1.5\times 10^{-4}$. |
| LRs | Differential: $\eta_{\text{CLIP}}$=$10^{-6}$; $\eta_{\text{heads}}$=$10^{-4}$; step ×0.1 @ {20,40}. |
| Sched/Run | Batch 32; up to 60 epochs; select by best *val accuracy*; AMP; grad clip (norm 5.0). |
| FT | Unfreeze only last $L_v$=4 visual blocks; text frozen ($L_t$=0); new modules end-to-end. |
| Loss | $\mathcal{L}$=0.5 $\mathcal{L}_g$ + 0.5 $\mathcal{L}_a$; gender CE (fused=1.0 + two auxiliaries=0.25 each); attrs CE with ignore $-100$; *Unknown* preserved. |
| Eval | Acc, Balanced Acc, F1 (macro), Precision (macro), Recall (weighted), AUC (Female vs Rest); ROC/CM per epoch. |

*Losses..* Total loss: $\mathcal{L}_{\text{total}} = 0.5\,\mathcal{L}_g + 0.5\,\mathcal{L}_a$. Gender loss uses categorical cross-entropy on the fused head (weight 1.0) plus two auxiliaries (weights 0.25 each). Attributes use per-attribute multiclass cross-entropy averaged over the five selected soft biometrics; missing labels use ignore index $-100$. The *Unknown* class is preserved to avoid forced binary decisions under ambiguity.

*Evaluation and logging..* Metrics per epoch: Accuracy, Balanced Accuracy, F1 and Precision (macro), Recall (weighted), and AUC (Female vs. Rest); we also store confusion matrices and ROC curves. The *best checkpoint* is selected by highest validation accuracy. Logs are saved as `.txt` and figures as `.png`. Reproducibility: fixed NumPy/PyTorch seeds, `num_workers`=8, `cudnn.benchmark=true`. The SCA module can be enabled per path (`use_sca_path1/2`) and remains fixed for the whole training run. .

Table 3: **Overview of evaluated models.**

| Model | Backbone | Observations |
|---|---|---|
| SNN-PAR [22] | ViT-B/16 [8] | Energy-efficient spiking baseline; learns multi-label embeddings and optimizes attribute–feature correspondence. |
| MambaPAR [17] | VisionMamba [30] | Mamba blocks for efficient long-range dependency modeling (spatial–semantic interaction). |
| LML [18] | Swin Transformer [27] | Label-balanced loss to mitigate prediction skew in multi-label PAR. |
| Rethinking [19] | Swin Transformer [27] | Unified training and identity-independent splits; revisits label dependencies. |
| ALM [20] | BN-Inception [28] | Weakly supervised multi-scale attention for attribute localization. |
| VAC [21] | ResNet-50 [29] | Attention consistency under geometric transforms to stabilize predictions. |
| APR Network [23] | ResNet-50 [29] | Multi-task (ID + attributes) with Attribute Re-weighting Module (ARM). |
| DeepMAR [26] | ResNet-50 [29] | Classical frequency-weighted sigmoid cross-entropy PAR baseline. |
| **Ours** | Dual CLIP [7] | Dual-path CLIP with attribute-conditioned fusion; gender + attribute supervision for long-range robustness. |

We optimize a weighted sum of gender and attribute losses, $\mathcal{L}_{\text{total}} = \lambda_g \mathcal{L}_g + \lambda_a \mathcal{L}_a$, with $\lambda_g = \lambda_a = 0.5$. For ternary gender (Male/Female/Unknown) we apply CE at the fused head (weight 1.0) and auxiliary CE at the direct and attribute-mediated heads (weights 0.25 each). For five attributes, we use per-attribute multiclass CE (ignore-index $-100$) and average across attributes.

### 4.3. Baselines and Configuration

*Comparative methods.*. We compare our method against widely cited PAR and multimodal/multi-task baselines spanning CNN/Transformer families, recent efficient/long-range designs, and imbalance-aware training. The goal is to isolate the gains from *semantic composition of attributes* and *CLIP-based fusion* under a *uniform* training/evaluation setup.

All methods follow the same pipeline as Sec. 4.2 (identical splits, resize/normalization, augmentation, optimizer, schedule). Where applicable, we keep input and batch size identical and report: *accuracy*, *balanced accuracy* (mA), *macro-F1/macro-precision*, *weighted recall*, and UC (Female vs. Rest). Table 3 lists models, backbones, and brief notes; method-specific knobs (e.g., unfreezing, prompt usage) follow standard reproducible practice to avoid asymmetric tuning.

### 4.4. Quantitative Results

We evaluate the proposed framework on the unified **U-DetAGReID** split using the official train/validation CSVs and the five long-range "champion" attributes (hairstyle, lower, feet, accessories, upper). Images are resized to 224×224 with CLIP normalization. Unless stated otherwise, training runs for 60 epochs with batch size 32, AdamW (weight decay $1.5 \times 10^{-4}$), and differential learning rates: $\eta_{\text{CLIP}} = 10^{-6}$ for unfrozen CLIP layers and $\eta_{\text{heads}} = 10^{-4}$ for projection/SCA/classifier heads; both decay by ×0.1 at epochs 20 and 40. We apply selective fine-tuning, unfreezing only the last $L_v$=4 visual blocks (text layers frozen), and optionally enable SCA per path. Model selection is done by best validation accuracy.

Table 4: Quantitative comparison on **U-DetAGReID** (validation split). All methods share the same protocol (224×224, 60 epochs, AdamW, official splits).

| **Model** | **Acc** (%) | **mA** (%) | **F1** (%) | **AUC** (%) |
|---|---|---|---|---|
| SNN-PAR [22] | 68.25 | 66.44 | 71.89 | 80.12 |
| MambaPAR [17] | 49.57 | 52.72 | 60.49 | 63.38 |
| LML [18] | 74.61 | 74.58 | 72.11 | 81.00 |
| Rethinking [19] | 76.01 | 76.39 | 71.66 | 86.00 |
| ALM [20] | 74.02 | 72.73 | 72.81 | 81.14 |
| VAC [21] | 69.89 | 73.86 | 71.82 | 77.32 |
| APR Network [23] | 71.25 | 71.02 | 72.44 | 81.48 |
| DeepMar [26] | 64.00 | 62.37 | 67.77 | 79.00 |
| Proposed Model | 76.25 | 75.72 | 75.77 | 84.37 |

Acc: accuracy; mA: balanced accuracy; F1 macro; AUC = Female vs. Rest (fused head).

Regarding gender, we considered three classes: (Male/Female/Unknown), we report from the fused head: Accuracy, Balanced Accuracy (mA), Macro-F1, Macro-Precision, Weighted Recall, AUC (Female vs. Rest), and per-epoch confusion matrices. Attribute supervision uses per-attribute cross-entropy with `ignore-index` $= -100$. The total objective is

$$\mathcal{L}_{\text{total}} = \alpha_1 \mathcal{L}_g + \alpha_2 \mathcal{L}_a,$$

where $\alpha_i$ specify the weights asociated to each term ($\alpha_i = 0.5$ in our experiments) and $\mathcal{L}_g$ includes the fused head (weight 1.0) and two auxiliary gender heads (direct and attribute-mediated; weights 0.25 each). All baselines and ablations share the same data pipeline and schedule; ablations vary SCA usage, the number of unfrozen visual blocks, and LR settings while keeping the attribute set and evaluation protocol fixed.

Table 4 shows that we attain the best macro-F1 (75.77%) among all compared methods and a competitive top-1 accuracy (76.25%), narrowly surpassing the strongest CNN/Transformer baselines (e.g., *Rethinking* at 76.01% Acc). The macro-F1 gains indicate that the dual-path fusion (direct visual + attribute-mediated) improves *class-wise balance*, particularly under long-range ambiguity where soft cues dominate. In contrast, *Rethinking* achieves a slightly higher balanced accuracy (mA) and AUC, suggesting that its calibration favors the minority/uncertain cases and the Female-vs-Rest ROC criterion. Overall, the proposed selective fine-tuning ($L_v$=4) with attribute supervision trades a small margin in mA/AUC for a consistent improvement in macro-F1, which is often more indicative of robustness under skewed evidence at standoff distances.

#### 4.4.1. Quantitative analysis by distance

We ana In each panel, we show the predicted probability distributions for *Female* (pink) and *Male* (blue) as smoothed densities, together with the decision threshold $\tau = 0.5$ (dashed). Shaded regions to the left/right denote false negatives for *Female* and false positives for *Male*. Separability is captured by the distance between peaks and the overlap between curves: less overlap $\Rightarrow$ higher TPR/TNR; more overlap $\Rightarrow$ more errors. As distance/height increases, the *Female* curve shifts left, reducing sensitivity (TPR), while the *Male* curve remains concentrated on the left, generally preserving specificity (TNR).

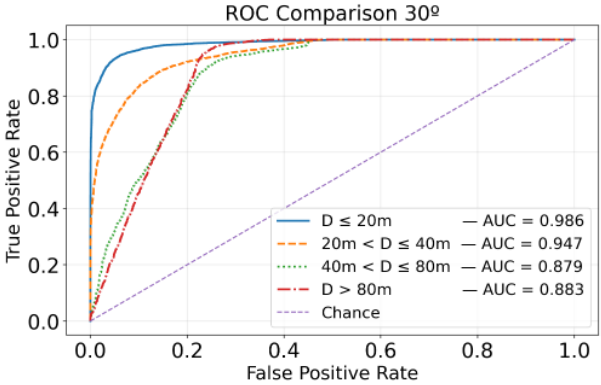

Figure 3: ROC Comparison at 30°. ROC curves across distance bins. AUC decreases progressively from short to long range: $D \leq 20$m (0.986), $20 < D \leq 40$m (0.947), $40 < D \leq 80$m (0.879), $D > 80$m (0.883).

*G1—Angle* 30° *(DetReIDx by distance)..* Across the four distance bins (Fig. 3), Female/Male score separability steadily decreases with distance. At *short range* ($D \leq 20$ m), discrimination is near perfect ($\mu_F$=0.929, $\mu_M$=0.079, $\sigma$=0.200), with TPR≈0.945 and TNR≈0.937 at $\tau$=0.5. At *near–medium* ($20<D\leq40$ m), separability declines ($\mu_F$=0.802, $\mu_M$=0.119, $\sigma$=0.271; TPR ≈ 0.834, TNR ≈ 0.899). For *medium–long* ($40<D\leq80$ m), overlap becomes substantial ($\mu_F$=0.674, $\mu_M$=0.181, $\sigma$=0.285; TPR ≈ 0.751). At *long* ($D$>80 m) there is a mild recovery ($\mu_F$=0.712, $\mu_M$=0.184, $\sigma$=0.264; TPR ≈ 0.836, TNR ≈ 0.797), yet Female remains left-shifted, indicating more false negatives. Overall, as resolution and soft-biometric cues degrade, Female scores drop faster than Male, narrowing the margin around the threshold.

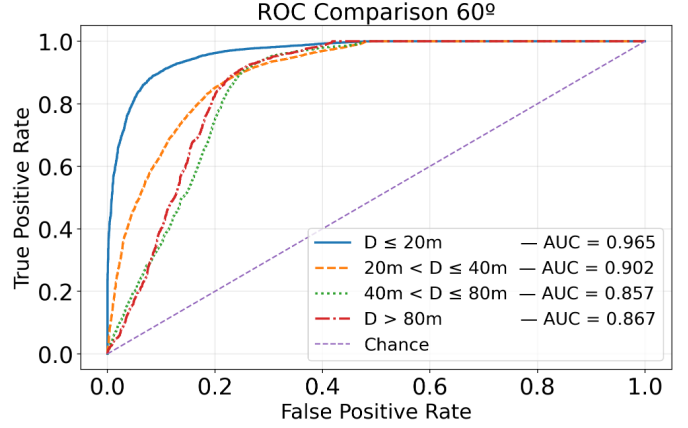

Figure 4: ROC Comparison at 60°. Similar trend as 30°, with faster degradation: AUC drops from 0.965 (short range) to 0.867 (long range).

*G1—Angle* 60° *(DetReIDx by distance)..* Degradation with distance is faster than at 30° (Fig. 4). At *short range* ($D \leq 20$ m), distributions remain well separated ($\mu_F$=0.86, $\mu_M$=0.10, $\sigma$≈0.25; TPR≈0.885, TNR≈0.918). At *near–medium* ($20<D\leq40$ m), overlap increases ($\mu_F$=0.71, $\mu_M$=0.16, $\sigma$≈0.29), reducing Female sensitivity (TPR≈0.753) while TNR≈0.856 holds. For *medium–long* ($40<D\leq80$ m), separability further declines ($\mu_F$=0.61, $\mu_M$=0.18, $\sigma$≈0.29; TPR≈0.651) with more FNs. At *long* ($D$>80 m), the pattern persists ($\mu_F$=0.63, $\mu_M$=0.18, $\sigma$≈0.28; TPR≈0.708, TNR≈0.828). Female shows a consistent leftward shift and higher uncertainty; Male remains comparatively stable.

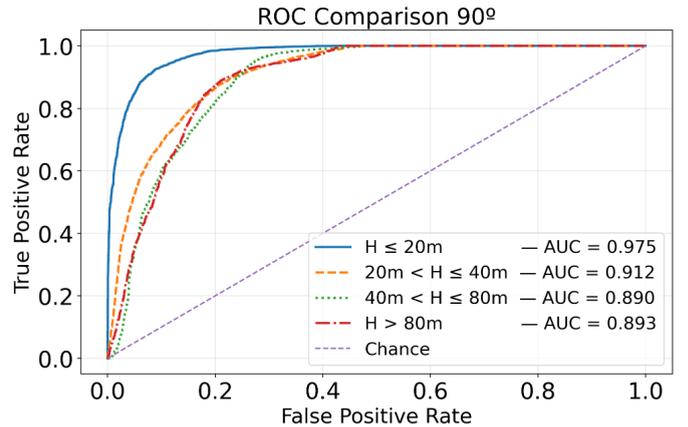

Figure 5: ROC Comparison at 90° (nadir). AUC decreases with altitude: $H \leq 20$m (0.975), $20 < H \leq 40$m (0.912), $40 < H \leq 80$m (0.890), $H > 80$m (0.893).

*G1—Angle* 90° *(DetReIDx by height)..* Under nadir view, class separability is strong at low heights but declines with $H$ (Fig. 5). At *low* ($H \leq 20$ m): $\mu_F$≈0.923, $\mu_M$≈0.129, $\sigma$≈0.231; TPR≈0.946, TNR≈0.882. At *near–medium* ($20<H\leq40$ m): $\mu_F$≈0.766, $\mu_M$≈0.176, $\sigma$≈0.290, with rising FNs and moderate operating rates (TPR≈0.817, TNR≈0.834). At *medium–long* (40<$H$≤80 m): $\mu_F$≈0.678, $\mu_M$≈0.165; TPR≈0.737, with growing overlap. At *very high* ($H$>80 m): $\mu_F$≈0.625, $\mu_M$≈0.148; TPR≈0.702, TNR≈0.863. Fine-grained cues fade with $H$, yielding Female underestimation while Male stays conservative.

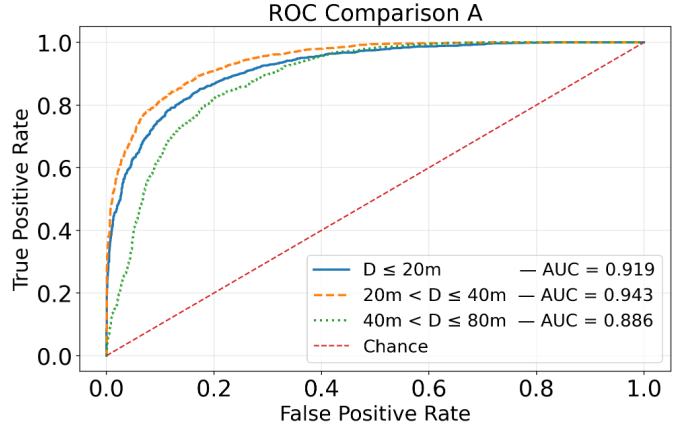

Figure 6: ROC Comparison on AG-ReID.v2. Strong sensitivity across tiers; AUC ranges from 0.919 (short) to 0.886 (mid). Long-range remains competitive.

*G2—AG-ReID.v2 (by distance tiers)..* Across tiers, Female means remain high ($\mu_F$≈0.87–0.89, $\sigma$≈0.26–0.30), while Male drifts right with distance ($\mu_M$≈0.26→0.39), raising FPs and lowering TNR. A0 ($D$≤20 m): $\mu_F$≈0.889, $\mu_M$≈0.264; TPR≈0.932, TNR≈0.761. A1 (20<$D$≤40 m): $\mu_F$≈0.889, $\mu_M$≈0.264; TPR≈0.932, TNR≈0.761. A2 ($D$>40 m): $\mu_F$≈0.887, $\mu_M$≈0.388; TPR≈0.947, TNR≈0.619. The proposed method generalizes well (stable $\mu_F$); however, specificity degrades at longer ranges (TNR: 0.76 → 0.76 → 0.62) as the Male distribution shifts rightward, increasing overlap with Female. A modest threshold shift (e.g., $\tau$≈0.55) or asymmetric costs can recover TNR in A2 with limited TPR loss.

*Table 5.* shows a consistent decline in class separability as distance (or camera height, for 90°) increases. For oblique views (30° and 60°), the model maintains strong discrimination at short ranges: for $D \leq 20$,m, Female and Male means are well separated (e.g., $\mu(F) \approx 0.93$ vs. $\mu(M) \approx 0.08$ at 30°) and operating points remain high (TPR/TNR ≈ 0.95/0.94 at 30° and ≈ 0.89/0.92 at 60°). Between 20 and 40,m, performance is still relatively strong, but beyond 40,m $\mu(F)$ decreases while $\mu(M)$ increases slightly, shrinking the margin between classes and increasing Female→Male errors; in parallel, Male specificity drops from about 0.94 to ≈ 0.80–0.83. In the nadir view (90°), performance also degrades with altitude: separability is high up to $H \leq 20$,m (TPR/TNR ≈ 0.95/0.88), but sensitivity drops to around 0.70 at $H > 80$,m, consistent with the loss of discriminative soft-biometric cues in near-vertical footage, while TNR remains in the 0.83–0.86

Table 5: **Summary of gender prediction distributions.** Means of $p_{\text{female}}$ for F/M and operating TPR/TNR. Positive class = Female.

| Group / Angle | Range | $\mu(\mathbf{F})$ | $\mu(\mathbf{M})$ | **TPR** | **TNR** |
|---|---|---|---|---|---|
| **G1 – 30°** | $D \leq 20$ m | 0.929 | 0.079 | 0.945 | 0.937 |
| | $20 < D \leq 40$ m | 0.802 | 0.119 | 0.834 | 0.899 |
| | $40 < D \leq 80$ m | 0.674 | 0.181 | 0.751 | 0.814 |
| | $D > 80$ m | 0.712 | 0.184 | 0.836 | 0.797 |
| **G1 – 60°** | $D \leq 20$ m | 0.861 | 0.102 | 0.885 | 0.918 |
| | $20 < D \leq 40$ m | 0.712 | 0.162 | 0.753 | 0.856 |
| | $40 < D \leq 80$ m | 0.607 | 0.182 | 0.651 | 0.821 |
| | $D > 80$ m | 0.631 | 0.179 | 0.708 | 0.828 |
| **G1 – 90°** | $H \leq 20$ m | 0.923 | 0.129 | 0.946 | 0.882 |
| | $20 < H \leq 40$ m | 0.766 | 0.176 | 0.817 | 0.834 |
| | $40 < H \leq 80$ m | 0.678 | 0.165 | 0.737 | 0.840 |
| | $H > 80$ m | 0.625 | 0.148 | 0.702 | 0.863 |
| **G2 — AG-ReID.v2** | A0 ($D \leq 20$ m) | 0.889 | 0.264 | 0.932 | 0.761 |
| | A1 ($20 < D \leq 40$ m) | 0.889 | 0.264 | 0.932 | 0.761 |
| | A2 ($D > 40$ m) | 0.887 | 0.388 | 0.947 | 0.619 |

range. For AG-ReID.v2 (G2), the model maintains high Female sensitivity across all tiers (TPR ≈ 0.93–0.95), but Male specificity declines with distance (TNR : 0.76 → 0.76 → 0.62), accompanied by an increase in $\mu(M)$ (from 0.26 to 0.39), indicating a rightward drift of the Male distribution and greater overlap between class scores at longer ranges.

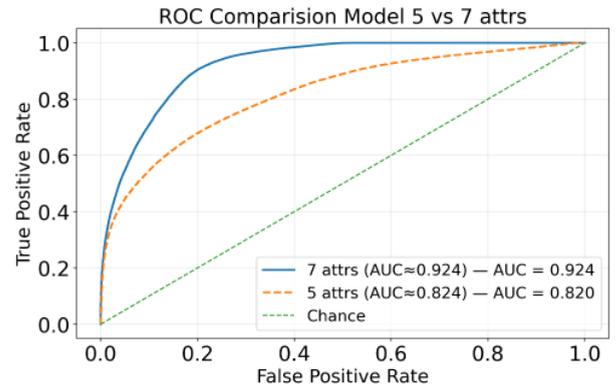

Figure 7: **ROC comparison between 5-attribute and 7-attribute models.** The 7-attribute model substantially outperforms the 5-attribute configuration (AUC = 0.820 vs. 0.924). The 7-attr ROC curve clearly dominates, with the largest gains at low false positive rates (FPR < 0.2), achieving higher TPR at matched FPR and improved class separability.

### 4.5. Qualitative Results

We qualitatively assess the proposed solution on **U-DetAGReID** across oblique (30°, 60°) and nadir (90°) views and varied ranges/altitudes ($D \in [10, 120]$ m, $H \in [5.8, 120]$ m). Each example shows fused confidence and distance tier ("D-xx"). We also visualize text→vision *attention rollout* [31] for key attributes (*hairstyle*, *upper*, *lower*, *feet*, *accessories*) and the fused gender head.

*Typical successes..* As in Fig. 8 (left), the model is reliable at 30°/60° up to ∼40 m and often remains confi-dent to ∼80 m. Correct cases align with clear, coherent cues (upper-color/shape, silhouette, skirt/pants patterns) despite missing faces. At 90°, accuracy is strong when torso/hip proportions are visible.

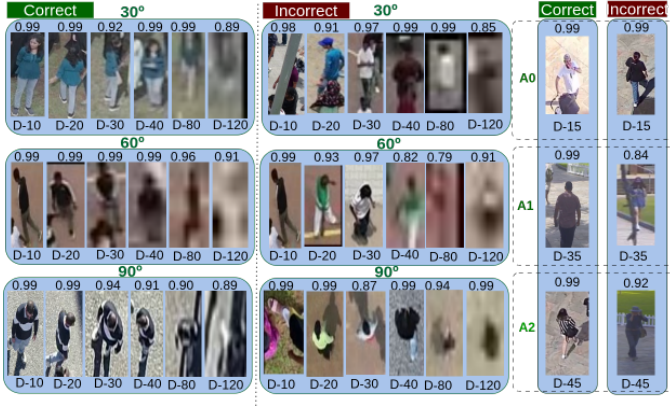

Figure 8: Correct vs. incorrect cases by viewing angle and distance. Each tile reports fused confidence and distance tier (D-xx). Left: correct predictions; right: failures under challenging regimes (distance/height/occlusion).

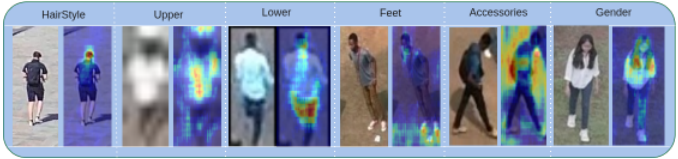

Figure 9: Attribute contribution maps via text-conditioned *Transformer attention rollout* [31]. Left→right: *Hairstyle*, *Upper*, *Lower*, *Feet*, *Accessories*, and *Gender* (fused). Warmer regions indicate stronger influence.

*Failure modes..* Fig. 8 (right) highlights: (1) Hair–upper fusion at 90°: dark hair/tops merge at $H \gtrsim 80$ m, inverting gender. (2) All-black at 120 m: boundaries vanish in nadir, inducing uncertainty. (3) Ambiguous silhouette at 80 m: wide gait/rectangular shoulders bias toward male when clothing cues are weak. (4) Shadows/ground artifacts: misattributed as subject at $D{\geq}80$ m or $H{\geq}80$ m. (5) Partial field of view: lower-only crops (shorts/sneakers) trigger Female→Male errors. (6) Multiple subjects: salient female foreground can dominate when the labeled person is background. (7) Label noise/ambiguity: heavily occluded or low-detail cases can be debatable.

*Attribute-based explainability..* Maps in Fig. 9 (even at small inputs, e.g., 270×250 px) show: Hairstyle/Upper focus head/torso when contrast/structure exists; Lower/Feet capture trousers/skirts/shoes and remain stable at long range; Accessories help when delineated, less so in nadir; the Gender (fused) map integrates all cues, with diffuse heat indicating structural uncertainty.

*Summary..* Qualitative evidence matches the quantitative trends: dual-path fusion (direct vision + attribute reasoning) improves robustness under weak signals, while text-guided cross-attention stabilizes inference without faces. Degradation follows increasing distance/elevation and nadir geometry, consistent with physical limits of long-range imagery.

### 4.6. Ablation Study

*Setup..* We systematically ablated our design choices on **U-DetAGReID** validation. All variants share the same pipeline (AdamW optimizer, step decay at epochs 20/40, 224×224 CLIP-normalized inputs, auxiliary gender + per-attribute multiclass loss with `ignore_index`). Varied components: (i) visual/attribute pathways; (ii) Spatial–Channel Attention (SCA) modules; (iii/iv) visual/textual transformer blocks unfrozen (FT-Vis/Txt); (v) learning rates (CLIP vs. heads); (vi) dropout in fusion head.

*Key findings..* (1) Backbone resolution: ViT-B/32→B/16 improves all metrics (mA: 63.81%→75.72%), confirming finer patches capture richer long-range details. (2) Selective fine-tuning: Unfreezing only last 4–5 visual blocks (FT-Vis=4–5) improves performance without catastrophic forgetting, sufficient for domain specialization. (3) Text freezing: FT-Txt=0 preserves semantic grounding while visual embeddings adapt (B/16.v1 vs. v2: freezing text yields better mA/F1). (4) SCA modules: Dual SCA improves feature refinement under oblique/low-resolution conditions via cross-dimensional gating. (5) LR partitioning: Stable configs use $1{\times}10^{-6}$ (CLIP) vs. $1{\times}10^{-4}$ (heads), preventing semantic drift. (6) Dropout: Increasing to 0.4 (B/16.v2) enhances generalization and reduces overconfidence on imbalanced classes. (7) Best trade-off: B/16.v3 achieves highest mA/AUC (77.96%/86.79%) but F1 drops (52.16%), indicating overfitting. B/16.v2 balances accuracy (75.72%), F1 (75.77%), AUC (84.37%) with stable convergence. (8) SOTA config: B/16.v4 adds 2 attributes (7 total: *beard*, *moustache*, *accessories*), differential LR (1e-6 vs. 1e-5), achieving mA: 85.61%, F1: 85.60%, AUC: 94.11%, surpassing all baselines via improved semantic diversity and regularization.

These results validate our design: controlled adaptation (higher-resolution patching, moderate visual unfreezing, frozen text, dual SCA, differentiated LRs) enables robust multimodal reasoning. Additional attributes provide redundancy when dominant cues (clothing, hairstyle) are ambiguous, confirming that multiple weak cues, coherently fused, outperform single signals under extreme viewing conditions.

### 5. Discussion

We analyze how attribute choice, architecture, and viewing regimes (distance, height, occlusions) affect performance. We quantify attribute contributions, discuss class balance and ablation-driven configuration changes, examine cross-attention interpretability, and conclude with key takeaways.

Table 6: Ablation on U-DetAGReID validation. All configurations share the same data pipeline, attribute supervision, and auxiliary gender heads. SCA-1/2 indicate Spatial–Channel Attention in Path 1/2. FT-Vis/Txt specify unfrozen transformer blocks.

| Config | +Attr +img | img | SCA-1 | SCA-2 | FT-Vis | FT-Txt | LR (CLIP) | LR (Heads) | Drop | Ep | mA (%) | F1 (%) | AUC (%) |
|---|---|---|---|---|---|---|---|---|---|---|---|---|---|
| ViT-B/32.v1 | ✓(14) | ✓ | ✗ | ✗ | 0 | 0 | ✗ | 1e-4 | 0.2 | 128 | 63.81 | 61.06 | 76.21 |
| ViT-B/32.v2 | ✓(5) | ✓ | ✗ | ✓ | all | 0 | 1e-4 | 1e-4 | ✗ | 64 | 72.69 | 48.68 | 82.61 |
| ViT-B/32.v3 | ✓(5) | ✗ | ✗ | ✓ | 0 | 0 | ✗ | 1e-4 | ✗ | 64 | 73.78 | 73.38 | 81.33 |
| ViT-B/32.v4 | ✓(5) | ✗ | ✗ | ✓ | all | all | 1e-4 | 1e-4 | 0.5 | 32 | 73.81 | 50.29 | 83.66 |
| ViT-B/16.v1 | ✓(5) | ✓ | ✓ | ✓ | 4 | 4 | 1e-5 | 1e-4 | 0.2 | 16 | 74.69 | 74.97 | 83.86 |
| ViT-B/16.v2 | ✓(5) | ✓ | ✓ | ✓ | 4 | 0 | 1e-6 | 1e-4 | 0.4 | 32 | 75.72 | **75.77** | 84.37 |
| ViT-B/16.v3 | ✓(5) | ✓ | ✓ | ✓ | 5 | 0 | 1e-6 | 1e-4 | 0.2 | 32 | **77.96** | 52.16 | **86.79** |
| **ViT-B/16.v4[†]** | ✓(7) | ✓ | ✓ | ✓ | 4 | 0 | 1e-6 | 1e-5 | 0.2 | 32 | **85.61** | **85.60** | **94.11** |
| ViT-L/14.v1 | ✓(7) | ✓ | ✓ | ✓ | 0 | 0 | ✗ | 1e-5 | 0.2 | 32 | **81.61** | 81.88 | **92.20** |
| ViT-L/14.v2 | ✓(7) | ✓ | ✓ | ✓ | 6 | 0 | 1e-7 | 1e-5 | 0.5 | 64 | 81.14 | 81.41 | 91.80 |

[a] **FT-Vis/Txt**: number of last visual/textual blocks unfrozen; 0 = frozen.
[b] **+Attr+img**: uses both attributes and image; parentheses show attribute count.
[c] **Ep**: training epochs.

## 5.1. Attribute-Level Contribution

The seven soft-biometric attributes (*beard, moustache, hairstyle, upper, lower, feet, accessories*) provide complementary cues when faces are unavailable. Figure ?? shows per-attribute class imbalance (notably in *hairstyle* and *upper*), yet the fused representation remains stable. *Upper*, *lower*, and *hairstyle* dominate in typical scenes; *beard/moustache* help at short/mid range; *feet/accessories* retain signal at long range due to higher visibility, ground-plane contrast, and robustness to downsampling and perspective.

## 5.2. Effect of Configuration Changes

Two findings recur across results and ablations. (1) With dual-stream CLIP (ViT-B/16) and cross-attention fusion, the five-attribute model already surpasses prior work, evidencing strong synergy between visual and attribute paths. (2) After inspecting attribute balance, we increased dropout (0.2→0.4), adjusted the CLIP/heads LR split, and expanded to seven attributes, yielding further gains and stability. Improvements arise from calibrated regularization plus added semantic diversity, not attribute count alone, and better align the model to long-range conditions.

## 5.3. Cross-Attention Behavior and Interpretability

Qualitative inspection (Sec. 4.5) shows text-conditioned cross-attention stabilizes predictions as resolution drops (e.g., $D>80$ m, 90° nadir). *Upper/lower* steer attention to body structure and clothing contrast; *feet* often preserve separability under occlusion; *accessories* provide high-contrast anchors when present. These maps make decisions auditable, revealing how multiple weak cues coherently combine when facial evidence is absent.

## 5.4. Overall Insights

(i) Soft-biometric reasoning plus visual embeddings is *necessary* to maintain accuracy and interpretability under distance/height/occlusions. (ii) The five-attribute model is lean and strong; the seven-attribute variant, with stronger regularization and tuned LRs, offers broader semantics and robustness with modest overhead. (iii) Results validate the key hypothesis behind this paper: *multiple details, coherently fused, yield a single reliable decision* in long-range scenarios.

## 6. Conclusion

In this paper, the main hypothesis is that inferring a single solution upon multiple other soft attributes enables long-range, face-independent gender recognition contributes for recognition effectiveness, when the data is of very poor quality. For such purpose, we fused direct visual and attribute-mediated paths via text→vision cross-attention, maintaining accuracy and interpretability under severe standoff conditions.

On **U-DetAGReID**, the 5-attribute proposed model outperforms all baselines (**Acc** 76.25%, **mA** 75.72%, **F1** 75.77%, **AUC** 84.37%). With class-balance analysis, regularization/LR tweaks, and 7 attributes: **mA 85.61%**, **F1 85.60%**, **AUC 94.11%**.

Stratification shows controlled degradation. At 30°: $\mu=0.883$ (88.91%>0.5) for $D\leq20$ m → $\mu=0.611$ (63.12%>0.5) for $D>80$ m. At 60°: $\mu=0.831$ (83.65%) → $\mu=0.580$ (59.35%). Nadir (90°): $\mu=0.866$ (87.17%) for $H\leq20$ m → $\mu=0.540$ (55.17%) for $H>80$ m. On AG-ReID.v2, curves remain robust (A2: $\mu=0.747$, 76.06%>0.5), indicating generalization.

Ablations confirm the recipe: ViT-B/16, unfreezing the last 4–5 visual blocks (FT-Txt= 0), dual SCA, and differential LRs ($1\times10^{-6}$ for CLIP vs. $1\times10^{-4}$ for heads) yield the best stability/adaptation. Variants high in mA/AUC but low in F1 expose imbalance, mitigated by stronger dropout and broader attribute coverage.

Qualitative maps show *beard/moustache* dominate at short/mid range; *feet/accessories* at long range; the *Unknown* class prevents overconfidence.

In short, the proposed solution achieves robust, auditable gender recognition at 80–100+ m without faces. Future work: refine nadir uncertainty, incorporate geometric/temporal cues (multi-frame), and expand context-aware prompts for physics-aligned multimodal reasoning.

**Acknowledgements**

This work is funded by national funds through FCT – Fundação para a Ciência e a Tecnologia, I.P., and, when eligible, co-funded by EU funds under project/support UID/50008/2025 – Instituto de Telecomunicações, with DOI identifier https://doi.org/10.54499/UID/50008/2025.